\documentclass[10pt,twocolumn,letterpaper]{article}

\usepackage{iccv}
\usepackage{times}
\usepackage{epsfig}
\usepackage{graphicx}
\usepackage{amsmath}
\usepackage{amssymb}
\DeclareMathOperator*{\argmin}{argmin}
\newcolumntype{P}[1]{>{\centering\arraybackslash}p{#1}}

\usepackage{booktabs, multirow} 
\usepackage{soul}
\usepackage[table]{xcolor} 
\usepackage{changepage,threeparttable} 

\usepackage[breaklinks=true,bookmarks=false]{hyperref}

\iccvfinalcopy 

\def\iccvPaperID{****} 

\ificcvfinal\fi

\begin{document}

\title{GeoSpark: Sparking up Point Cloud Segmentation with Geometry Clue}

\author{Zhening Huang$^{1}$ \hspace{1.0cm} Xiaoyang Wu$^{2}$\hspace{1.0cm}  Hengshuang Zhao$^{2}$\hspace{1.0cm} Lei Zhu$^{3}$\hspace{1.0cm} Shujun Wang$^{1}$ \\ Georgios Hadjidemetriou$^{1}$ \hspace{1.0cm} Ioannis Brilakis$^{1}$ \\
\vspace{1.0cm}
$^{1}$University of Cambridge~~~
$^{2}$The Unversity of Hong Kong~~~
$^{3}$HKUST(GuangZhou)~~~
}

\maketitle
\ificcvfinal\thispagestyle{empty}\fi

\begin{abstract}
Current point cloud segmentation architectures suffer from limited long-range feature modeling, as they mostly rely on aggregating information with local neighborhoods.
Furthermore, in order to learn point features at multiple scales, most methods utilize a data-agnostic sampling approach to decrease the number of points after each stage. Such sampling methods, however, often discard points for small objects in the early stages, leading to inadequate feature learning.
We believe these issues can be mitigated by introducing explicit geometry clues as guidance.
To this end, we propose \textbf{GeoSpark}, a Plug-in module that incorporates \textbf{Geo}metry clue into the network to \textbf{Spark} up feature learning and downsampling. GeoSpark can be easily integrated into various backbones. For feature aggregation, it improves feature modeling by allowing the network to learn from both local points and neighboring geometry partitions, resulting in an enlarged data-tailored receptive field. Additionally, GeoSpark utilizes geometry partition information to guide the downsampling process, where points with unique features are preserved while redundant points are fused, resulting in better preservation of key points throughout the network. 
We observed consistent improvements after adding GeoSpark to various backbones including PointNet++, KPConv, and PointTransformer. Notably, when integrated with Point Transformer, our GeoSpark module achieves a 74.7\% mIoU on the ScanNetv2 dataset (4.1\% improvement) and 71.5\% mIoU on the S3DIS Area 5 dataset (1.1\% improvement), ranking top on both benchmarks. 
Code and models will be made publicly available.
\end{abstract}

\section{Introduction} 

\begin{figure}[t!]
 \centering
 \includegraphics[width=\linewidth]{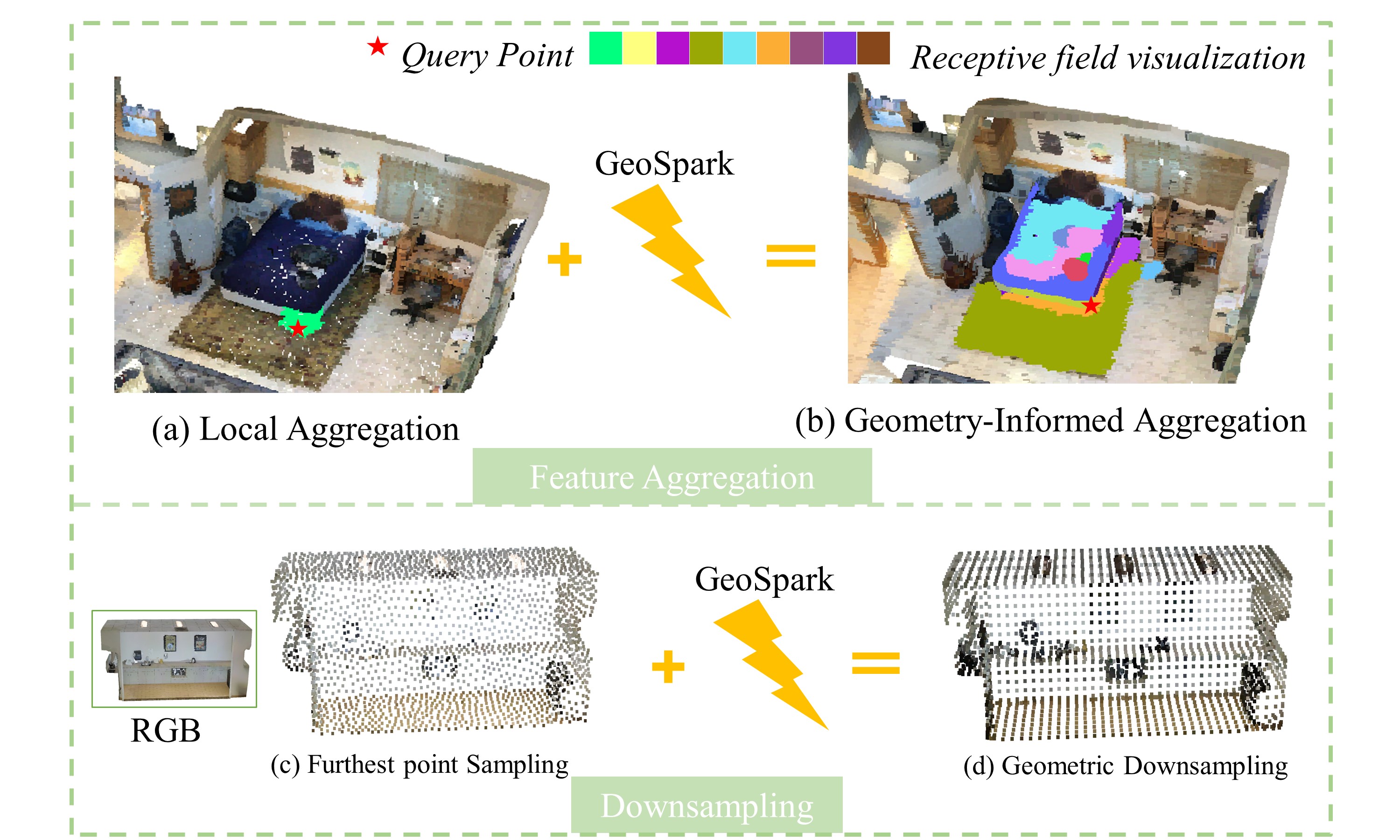}
 \caption{\textbf{Illustration of \textbf{GeoSpark} on feature aggregation and downsampling modules.} Current feature learning predominantly depends on local aggregation (shown in \textbf{a}). Introducing GeoSpark not only expands the receptive field but also concentrates on the relevant areas (shown in \textbf{b}). For instance, if the query point is located at the end of the bed, Geometry-Informed Aggregation will expand the receptive field to cover the entire area of the bed, as well as the bed mat on the floor. Moreover, \textbf{GeoSpark} also enhances sampling by using geometry clues as guidance. Compared to FPS (shown in \textbf{c}), Geometric Downsampling (shown in \textbf{d}) samples coarsely in areas of simple geometry and densely in areas with complex geometry, such as a dishwasher, a painting on the wall, and objects on the table. The joint of these two modules yields impressive segmentation results.} 
 \label{stitching_effect}
 \vspace{-4mm}
\end{figure}

Since PointNet++\cite{Qi2017PointNet++:Space}, mainstream point cloud segmentation methods \cite{Zhao2019Pointweb:Processing, Jiang2018PointSIFT:Segmentation, LiPointCNN:Points, Boulch_2020, Thomas2019KPConv:Cloudsb} conduct feature aggregation with local neighborhood, because learning global features for large-scale point clouds is infeasible. However, due to the redundancy in point clouds, local neighborhoods often contain a high percentage of similar points, limiting models' ability to learn from diverse contexts. For this reason, how to model long-range features in point cloud is a long-standing challenge in point cloud processing \cite{Lai_2022, Hu2019RandLA-Net:Clouds, Peng2023}.

Moreover, hierarchical paradigms are predominantly utilized in point cloud understanding architectures, where points undergo multiple downsampling stages to learn features at various scales. These frameworks typically employ data-agnostic sampling methods, such as FPS \cite{Qi2017PointNet++:Space}, voxelization \cite{Graham_2015}, or random sampling \cite{Hu2019RandLA-Net:Clouds}, which tend to dilute points for small objects. This is due to the severe data imbalance issues present in point clouds. The reduction in the number of points during later stages restricts the model's ability to learn expressive features for small objects, resulting in unsatisfactory performance.

 \begin{figure*}[ht!]
 \centering
 \includegraphics[width=1.0\textwidth]{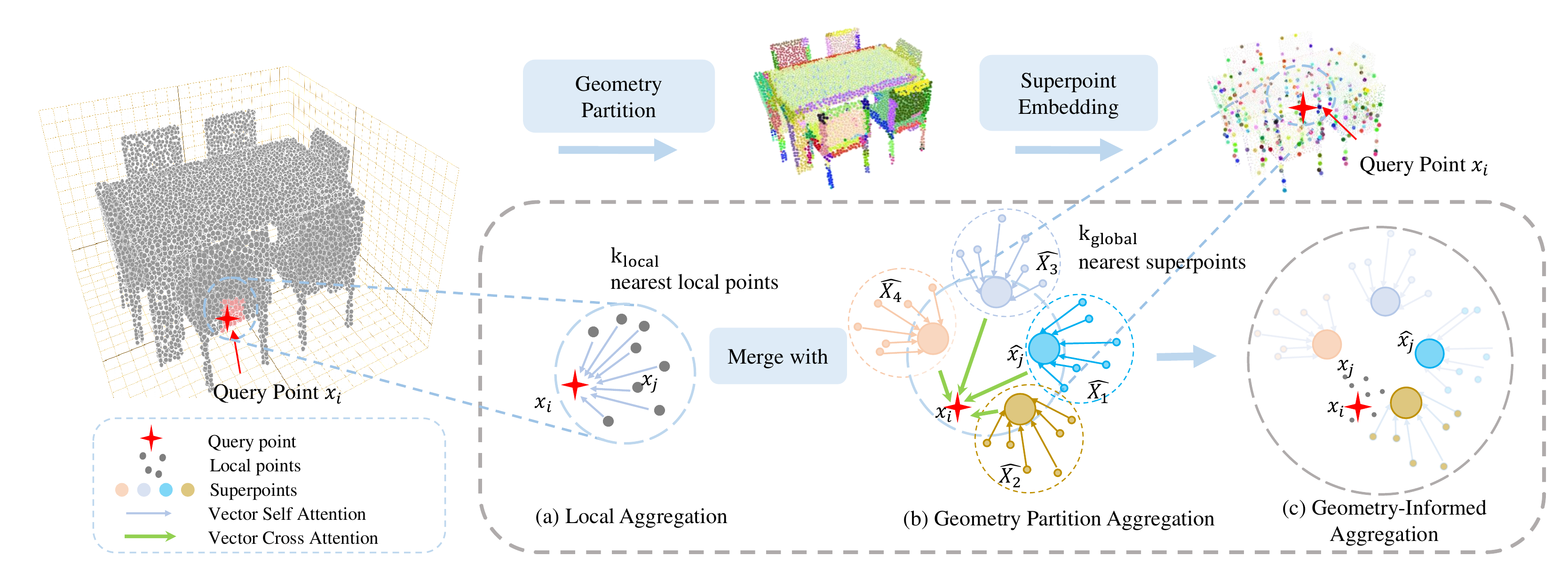}
 \centering
\vspace{-3mm}
  \caption{Illustration of \textit{Geometric Partition \& Embedding} and \textit{Geometry-Informed Aggregation} Modules. Geometric Partition \& Embedding split point clouds into simple geometric shapes, such as table surfaces, chair backs, table edges etc, which are then embedded into superpoints. The query point \(x_i\) conducts local aggregation with \(k_{local}\) local neighbour points \(x_j\), and then performs Geometry Partition aggregation with \(k_{global}\) superpoints \(\hat{x_j}\) to acquire global contexts. The \textit{Geometry-Informed Aggregation} module allows the model to learn both local fine details as well as expressive global contexts.}
 \label{attention_ill}
 \vspace{-4mm}
\end{figure*}

We argue that these challenges can be mitigated by introducing explicate geometry clue into the framework and preformed geometry-aware feature learning and downsampling. We define the geometry information as an initial geometry partition that can be obtained with conventional point cloud processing techniques \cite{Hui_2021, LandrieuCutFunctions, Landrieu2019PointLearning,Guinard2017WeaklyClouds,Lin_2018} where redundancy point clouds can be clustered into simple shapes. To this end, we introduce a Plug-in module named {\textbf{GeoSpark}} that can be integrated with various point cloud segmentation backbones to spark up semantic segmentation results by enhancing feature aggregation and the downsampling process. 

Specifically, for feature aggregation, GeoSpark offers a \textit{Geometry-Informed Aggregation} (GIA) module which attends to both local neighbor points and geometry partition entities. Unlike methods that rely solely on local neighborhoods, our approach offers two compelling advantages. Firstly, it significantly enlarges the receptive field of the feature aggregation module without dramatically increasing the number of reference points. This is achieved by encoding geometry partition into superpoints, which provide a compact yet information-rich format for the raw input. Secondly, the geometry partition entities are highly tailored to the input point cloud data. Consequently, the receptive field for each point is tailored to its specific geometry, enabling the model to learn from relevant regions rather than data-agnostic areas. This concept is akin to the methodology employed in the deformable structure \cite{DBLP:journals/corr/abs-2010-04159, Xia_2022_CVPR, Thomas2019KPConv:Clouds, DBLP:journals/corr/DaiQXLZHW17}, which learns offset values from the originally prescribed regions. Here, we achieve a similar effect without requiring additional offset design.

For the downsampling module, GeoSpark introduces \textit{Geometric Downsampling} (GD). Unlike conventional methods such as Further Point Sampling, Random Sampling, and Voxelization which drop points without considering their importance, our approach uses geometry partition as guidance and fuses points of redundancy into one while keeping points with unique features. In other words, the downsampling is the process of reducing point redundancy, ensuring that important points are retained. This data-dependent downsampling approach has been shown to be very beneficial, especially for small objects.

These two simple yet powerful module designs can be incorporated into various backbone models. We tested them on three backbone models, including Pointnet++ \cite{Qi2017PointNet++:Space}, KPConv \cite{Thomas2019KPConv:Cloudsb}, and Point Transformer \cite{Engel_2021}, and consistently observed improvements. In summary, our contribution is threefold.

\begin{itemize}
\item We demonstrate that utilizing geometric partition to enhance feature aggregation is an effective approach for performance gain. It not only expands the receptive field in a cost-efficient manner but also concentrates on relevant regions that are customized for the query point.

\item We enhance the downsampling module in the architecture by fusing redundant points and retaining points with unique features, resulting in better information retention throughout the network.

\item GeoSpark has proven to be universal and transferable to various baseline models. Notably, when incorporating it with Point Transformer \cite{Zhao_2021}, it yielded a 4.1\% improvement in the mIoU score on the ScanNetV2 \cite{Dai_2017} dataset, and achieved a score of 71.5\% on the S3DIS \cite{Armeni_2016} dataset, ranking high on both benchmarks.

\end{itemize}

\section{Related work} \label{sec:intro}
\paragraph{Point Cloud Segmentation.}
Deep learning methods for point cloud processing can be categorized into three types: projection-based methods \cite{Tatarchenko2018Tangent3D, Wu_2019,Wu_2018, Boulch_2018, Jaritz_2019}, voxel-based methods \cite{Choy20194DNetworks,Graham_2015, Park_2022}, and point-based methods \cite{Charles_2017,Qi2017PointNet++:Space, Zhao2019Pointweb:Processing, Hu2019RandLA-Net:Cloudsb, Jiang2018PointSIFT:Segmentation, Li2018, Boulch_2020}. Recent work on point-based methods mostly adopts an encoder-decoder paradigm while applying various sub-sampling methods to expand the receptive field for point features. A variety of local aggregation modules were developed, including convolution-based methods \cite{Thomas2019KPConv:Cloudsb, Boulch_2020}, graph-based methods \cite{Landrieu2018Large-scaleGraphs, Landrieu2019PointLearning}, and attention-based methods \cite{Hu2019RandLA-Net:Clouds, Zhao_2021}. Meanwhile, different downsampling methods \cite{Chen2022SASA:Detection, Dovrat_2019, Yan2020PointASNL:Sampling}, upsampling \cite{Qiu2021SemanticFusion, Varney_2022}, and post-processing methods \cite{Hu_2020, Lu_2021} have been proposed to enhance the network's robustness. The inherent permutation invariance of attention operations makes them well-suited for point cloud learning, leading to the application of transformer operations in point clouds \cite{Zhao_2021, Engel_2021, Xie_2020}, following their success in 2D and NLP. However, global self-attention is impractical due to massive computational costs, leading to recent work utilizing local-scale self-attention to learn point features. Point Transformer \cite{Zhao_2021} proposed a vector self-attention with k nearest neighbor points for feature aggregation and introduced the concept of learnable positional embedding, which has proved to be very powerful across various tasks. Nevertheless, local attention has a limited receptive field and cannot explicitly model long-range dependencies. Thus, Stratified Transformer \cite{Lai_2022} was developed and adopted a shifted window strategy to include long-range contexts. Similarly, Fast Point Transformer \cite{Park_2022} utilized a voxel hashing-based architecture to enhance computational efficiency. However, these data-agnostic methods only select receptive fields based on spatial information, which could fail to focus on relevant areas.
\vspace{-5mm}
\paragraph{Geometric Partition}
Geometric partitions are an over-segmentation technique for point clouds that groups point with similar geometric features. Over the years, several methods have been developed to learn hand-crafted features of point clouds and utilize clustering or graph partition methods to generate meaningful partitions \cite{Papon_2013, Guinard2017WeaklyClouds, Lin_2018}. However, the performance of these methods is limited by the handcrafted feature of point clouds. SPNet \cite{Hui_2021} introduced a superpoint center updating scheme that generates superpoints with the supervision of point cloud labels. Although very efficient, SPNet fails to produce intuitive partitions beyond grouping points with the same semantic label together. On the other hand, the supervised superpoint method uses MLP to learn point features and combines points with graph-structured deep metric learning \cite{Landrieu2018Large-scaleGraphs}. Nevertheless, Geometric partition techniques offer impressive outcomes in finding geometric shapes in point clouds. However, despite their impressive performance, there have been few attempts to utilize these techniques in the deep learning framework \cite{Landrieu2018Large-scaleGraphs, Landrieu2019PointLearning}. In this regard, we propose GeoSpark, which leverages the geometry information stored in the partition to enhance feature aggregation and downsampling.

\section{GeoSpark}
Our goal is to leverage the explicit geometry clue stored in the geometric partitions to improve the feature aggregation and downsampling process. In this section, we will elaborate on these are achieved on GeoSpark. We will begin by explaining how the geometric partition is generated and embedding, followed by technical details of our \textit{Geometry-Informed Aggregation} (GIA) module as well as \textit{Geometric Downsampling} (GD) module. 

\begin{figure*}[ht]
 \centering
 \includegraphics[width=0.9\textwidth]{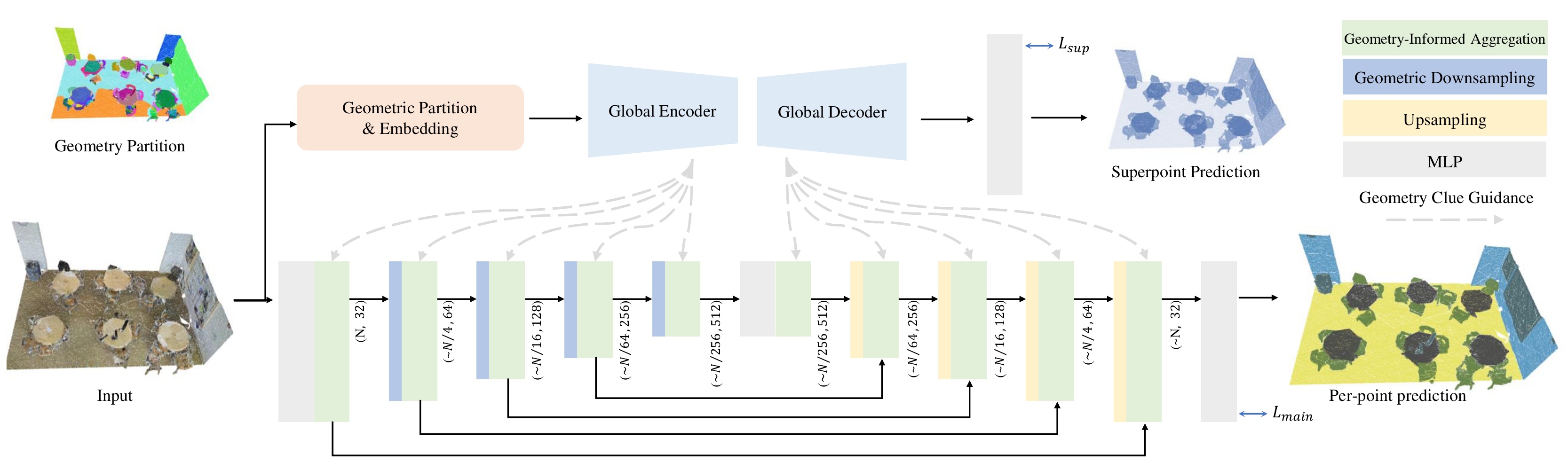} 
 \centering
 \caption{\textbf{The Paradigm of GeoSpark Architecture} Initially, the point input is fed into the "Geometric Partition \& Embedding" module to obtain geometric partitions and encode them into superpoints. These superpoints are then processed in the global branch to learn features at multiple scales before being added to the local branch as geometry guidance. In addition to the primary prediction loss, a superpoint loss is introduced to guide the training of the global branch. The aggregation module can be altered to various backbones.}

 \label{fig:networkTeaser}
\end{figure*}
\subsection{Geometric Partition \& Embedding}
The first step of the proposed GeoSpark network is to effectively learn the low-level features of point clouds and group similar points. This process significantly reduces the redundancy of point clouds and makes it possible to encode long-range context inexpensively in future steps. This module will process the entire input scene and produce an initial geometric partition, so it needs to be fast and effective. In our experiment, we tested different geometric partition methods, including VCCS\cite{Papon_2013}, Global Energy Model (GEM)\cite{Guinard2017WeaklyClouds}, Graph-Structured Method (GSM) \cite{Landrieu2019PointLearning} and SPnet \cite{Hui_2021}. The GEM \cite{Guinard2017WeaklyClouds} offers the best trade-off between speed and performance, so we integrate that into our \textit{Geometric Partition \& Embedding} module. Specifically, for each point, a set of geometric features, such as linearity, planarity, scattering, and verticality \(f_i \in \mathbb{R}^{c}\) are computed, characterizing local features and shapes. The geometrically homogeneous partition is defined as the constantly connected components and therefore as the solution to the following optimization problem\cite{Guinard2017WeaklyClouds}: 
\begin{equation}
 \argmin_{g \in \mathbb{R}^{c}} \sum_{i \in C} ||g_i-f_i||^2 + \lambda \sum_{(i,j)\in E_{nn}} \omega_{i, j}[g_i \neq g_j]
 \label{optimization}
 \vspace{-2mm}
\end{equation}
where \(\lambda\) is the \textit{regularization strength} that determines the coarseness of the partition, \(\omega_{i, j}\) is a weight matrix that is linearly inversely proportional to the distance between points and \([\cdot]\) is the Iverson bracket. We use the \textit{\(l_o\)-cut} pursuit algorithm\cite{LandrieuCutFunctions} to quickly find approximate solutions for Eq. \ref{optimization}, as it is impractical to find exact solutions when the number of points is really large. 
Given point cloud set \( X = (P, F)\), the geometric partition module splits the point cloud into point subsets \([\hat{X_1}, \hat{X_2}, \hat{X_3} \dots, \hat{X_m}] \), and each set contains a different number of points. 

To encode each partition into a superpoint, we adopted a lightweight MLP structure, inspired by PointNet \cite{Charles_2017}. The input point feature firstly go through MLP layers to gradually project the features of the points to higher dimension space. Points in each partition are then fused into one superpoint by applying MaxPooling in feature space \(\hat{f_i}\) and AvgPooling in coordinates space \(\hat{p_i}\). We concatenate global information of each partition \(\hat{f_{i,g}}\), such as \textit{partition diameter} into the features of the fused points, before applying other layers of MLP to reduce the feature dimension to \(c\). Formally,
\begin{equation}
\begin{aligned}
 \hat{F_i} &= T_2 \ ( \text{MaxPool}\ \{{T_1\ (\hat{f_i}) \ | \ \hat{f_i}\in \hat{X_i}}\} \ \oplus \ \hat{f_{i,g}} )\\
 \hat{P_i} &= \text{AvgPool} \ \{{\hat{p_i} \ | \ \hat{p_i}\in X_i}\} 
\end{aligned}
\end{equation}
where \(T_1 : \mathbb{R}^c \xrightarrow{} \mathbb{R}^{c_{l}}\) and \(T_2 :\mathbb{R}^{c_l} \xrightarrow{} \mathbb{R}^c\) are MLP layers that enlarge and condense point feature dimensions. 

\subsection{Geometry-Informed Aggregation}
The key idea of Geometry-Informed Aggregation (GIA) is to attend point feature both from local neighbour points and global geometric partition regions, encoded in superpoints. Therefore, GIA module takes two sets of inputs: local points \( X\ = (p_i, f_i)\) and encoded superpoints \(\hat{X} =(\hat{p_i}, \hat{f_i})\). Local point features \(f_i\) and superpoint features \(\hat{f_i}\) are first processed by separated MLP layers before being fed into the \textit{GIA} module. The following explain the proposed method using Point Transformer as backbone. However, similar strategies can be incorporated with other backbone easily.

\paragraph{Local Neighbor Aggregation}
The local neighbor aggregation follows the orginal design of backbone structure. For Point Transformer, it \cite{Zhao_2021} first uses three linear projections \(\phi, \psi, \alpha\) to obtain \textit{query}, \textit{key} and \textit{value} for local points. Following \textit{vector self-attention}, a weight encoding function \(\omega\) is introduced to encode the subtraction relations between point \textit{queries} and \textit{keys}, before processing to the \textit{softmax} operation to form the attention map. A trainable parameterized position encoding, formed by an MLP encoding function \(\theta\), is added to both the attention map and the value vectors. Formally, for query point \(x_i = (p_i, f_i)\) and reference points \(x_j= (p_j,f_j)\), the local attention map is formed as follows:
\begin{equation}
\begin{aligned}
 \delta_{i,j} & = \theta(p_i-p_j) \\
w_{i,j} &= \omega(\phi(f_{j})- \psi(f_{i})+\delta_{i,j}) \\
\end{aligned}
\end{equation}
\paragraph{Geometric Partition Aggregation}

To learn point feature from superpoint, a feature map is built between local points and global superpoints. Specifically, \textit{key} and \textit{value} is projected from superpoint features \(\hat{f_i}\) with linear operation \(\Psi\), and \(A\) respectively. Attention is formed similarly to \textit{Local Attention}. However, an independent weight encoding function \(\Omega\) is used and the positional embedding also employs a different MLP \(\Theta\) to encode the coordinate difference between the query points and the reference superpoints. 
\begin{equation}
\begin{aligned}
\Delta_{i,j} &= \Theta(p_i-\hat{p_j}) \\
W_{i,j} &= \Omega(\phi(\hat{f_{i}})- \hat{\Psi(f_{j})} +\Delta_{i,j})\\
\end{aligned}
\end{equation}
In practice, \textit{Local Neighbor Aggregation} and \textit{Geometric Partition Aggregation} are performed simultaneously and merged to form \textit{Geometry-Informed Aggregation}, formally:
\begin{equation}
\begin{aligned}
f_{i,local}' &= \sum_{f_i \in K} \text{softmax}(w_{i,j}) \odot (\alpha(f_j)+\delta_{i,j}) \\
f_{i,global}' &= \sum_{\hat{f_i} \in \hat{K}}\text{softmax}(W_{i,j}) \odot (A(\hat{f_j})+\Delta_{i,j})\\
f_i^{'} &= \xi \left[{f_{i,local}' + f_{i,global}'}\right]
\end{aligned}
\end{equation}
where \(\odot\) represents Hadamard product, \(K\) is defined as reference local points, taken from \(k_{local}\) nearest neighbour, and \(\hat{K}\) is the reference superpoints, sampled with \(k_{global}\) nearest neighbour. After merging the outputs of \textit{Local Neighbor Aggregation} and \textit{Geometric Partition Aggregation}, an MLP layer (\(\xi : \mathbb{R}^c \xrightarrow{} \mathbb{R}^{c}\)) is employed to further glue global and local features and form the final outputs. Notice that it is important to maintain a good global and local balance to ensure that both fine details and long-range dependencies are included. More details on this can be found in the ablation study.

The features of superpoints \(\hat{X} =(\hat{p_i}, \hat{f_i})\) are learned in a separate global branch. In this branch, the superpoint features are projected to different dimensions at various stages to match the dimension space of the local branch. Since the number of superpoints is typically several orders of magnitude smaller than that of the local branch, the feature learning in the global branch is very fast.

\paragraph{Loss function.} A simple yet important superpoint loss is introduced to assist feature learning in the global branch. 
Specifically, given a local point set \(X\), and its label \(E = \{e_i \in \mathbb{R}^l \ | \ i=1,\dots, n\}\), where \(e_i\) is the one-hot vector. We generate a \textit{soft pseudo label} by calculating the label distribution in each geometric partition \(W = \{w_j \in \mathbb{R}^l\ | \ j = 1, \dots, m\} \). We optimize the global branch by minimising the distance between superpoint prediction \(u\) and \textit{soft pseudo label} \(w\). Overall, our loss is constructed with per-point prediction loss and the superpoint loss while a parameter \(\beta\) is introduced to determine the weight of the superpoint loss. Given the pre-point predication as \(e'\), the loss function is
\begin{equation}
 L_{total} = \frac{1}{n}\sum_{i=1}^{n}\mathcal{L}_{loss}(e_i,e_i') + \beta \frac{1}{m}\sum_{i=1}^{n}\mathcal{L}_{loss}(w_j,u_j)
\end{equation}
where \(\mathcal{L}_{loss}\) can be various type of loss function. Cross-entropy is selected in our experiments.

\subsection{Geometric Downsampling}
\begin{figure}[b!]
 \centering
 \includegraphics[width=0.5\textwidth]{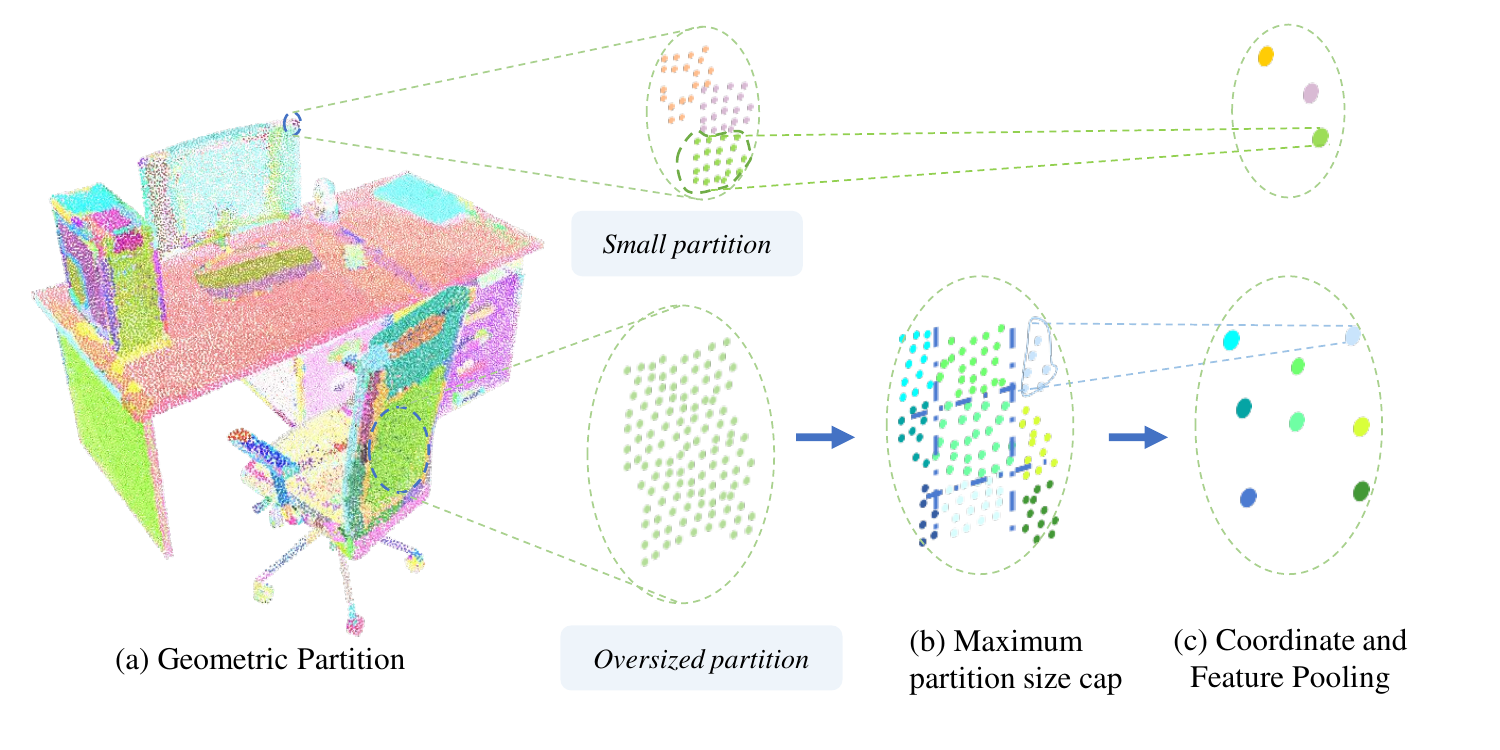}
 \centering
 \caption{\textbf{Geometric Downsampling.}
\textit{Geometric Downsampling} module fuses points in small partitions into superpoints and further split oversized partitions with a grid to control the downsampling rate. This method drops redundant points and retains points with unique features}
\label{fig:GD}
\end{figure}

One notable challenge in the downsampling process is the early loss of the under-representative points, such as points for small objects, leading to insufficient feature learning and unsatisfactory prediction results. Data agnostic sampling methods such as random sampling, and FPS, do not consider the uniqueness of points and are likely to cause this issue. To mitigate this problem, we develop the \textit{Geometric Downsampling} module with the help of geometric partitioning.
The motivation is to preserve points with unique features and drop redundancy points in the downsampling process. Given a point set \( X = \{(p_i, f_i)\ |\ i=1 \dots n\}\), instead of sampling from the whole point set, we conduct sampling from each geometric partition. In particular, we set a target diameter \(a\) for new points, and further split every partition that is larger than the predetermined size \(a\) with voxel grids to obtain subsets, e.g. \([\hat{X_1}]\) to \([\hat{X_{11}}, \hat{X_{12}}, \hat{X_{13}}, \dots]\). Once this process is done, we fuse the fine partition with pooling operations, where \textit{MaxPooling} is applied at features space and \textit{AvgPooling} is conducted on coordinate space, as illustrated in Figure \ref{fig:GD}. Similar to Point Transformer, we apply MLP layers \(D: \mathbb{R}^c \xrightarrow{} \mathbb{R}^{c}\) on feature space before pooling. Mathematically: 
\begin{equation}
\begin{aligned}
{f_{2,i}} = \text{MaxPool}\{Df_{1, i}|f_i\in X_{1, ii}\}\\
{p_{2,i}} = \text{AvgPool}\{p_{1, i}|p_i\in X_{1, ii}\} 
\end{aligned}
\end{equation}

\section{Experiments}
We evaluated the effectiveness of GeoSpark by integrating it with three backbones: PointNet++ \cite{Qi2017PointNet++:Space}, KPConv \cite{Thomas2019KPConv:Clouds}, and PointTransformer \cite{Zhao_2021}. We selected two competitive indoor datasets for testing, including the Stanford 3D Indoor dataset (S3DIS) \cite{Armeni_2016} and ScanNetV2 \cite{Dai_2017}. We compared the results of GeoSpark with the corresponding baselines and other state-of-the-art methods, as shown in Table \ref{s3dis_pre_class} and Table \ref{scanv2}.

We also conducted comprehensive ablation studies to evaluate various design decisions, which involved analyzing individual modules (refer to Table \ref{MODULE}), local-global balance for GIA (refer to Table \ref{balance}), sampling strategies (refer to Table \ref{ab_samp}), and the geometric partition size (refer to Table \ref{partition_cap}).

\subsection{Experiments Setting}
\paragraph{Network architecture.} 

\begin{table*}[!th]
	\centering
            \resizebox{\textwidth}{!}{
		\begin{tabular}{ l |l l | c c c c c c c c c c c c c}
			\toprule[1pt]
			Method  & mAcc (\%) & mIoU (\%) & ceiling & floor & wall & beam & column & window & door & table & chair & sofa & bookcase & board & clutter \\
			\hline
PointNet\cite{Charles_2017} &49.0 &41.1 &88.8 &97.3 &69.8 &0.1 &3.9 &46.3 &10.8 &59.0 &52.6 &5.9 &40.3 &26.4 &33.2 \\
{SegCloud\cite{Tchapmi_2017}}  &57.4 &48.9 &90.1 &96.1 &69.9 &0.0 &18.4 &38.4 &23.1 &70.4 &75.9 &40.9 &58.4 &13.0 &41.6 \\
{TangentConv\cite{Tatarchenko2018Tangent3D}}  &62.2 &52.6 &90.5 &97.7 &74.0 &0.0 &20.7 &39.0 &31.3 &77.5 &69.4 &57.3 &38.5 &48.8 &39.8 \\
{PointCNN \cite{LiPointCNN:Points}}  &63.9 &57.3 &92.3 &98.2 &79.4 &0.0 &17.6 &22.8 &62.1 &74.4 &80.6 &31.7 &66.7 &62.1 &56.7 \\
{SPGraph \cite{Landrieu2018Large-scaleGraphs} } &66.5 &58.0 &89.4 &96.9 &78.1 &0.0 &42.8 &48.9 &61.6 &84.7 &75.4 &69.8 &52.6 &2.1 &52.2 \\
{PCCN \cite{Wang2021DeepNetworks}}&67.0 &58.3 &92.3 &96.2 &75.9 &0.3 &6.0 &\textbf{69.5} &63.5 &66.9 &65.6 &47.3 &68.9 &59.1 &46.2 \\
{PAT \cite{Yang2019ModelingSampling}}  &70.8 &60.1 &93.0 &98.5 &72.3 &\textbf{1.0} &41.5 &85.1 &38.2 &57.7 &83.6 &48.1 &67.0 &61.3 &33.6 \\
{PointWeb\cite{Zhao2019Pointweb:Processing}} &66.6 &60.3 &92.0 &98.5 &79.4 &0.0 &21.1 &59.7 &34.8 &76.3 &88.3 &46.9 &69.3 &64.9 &52.5 \\
{HPEIN\cite{Jiang2019HierarchicalSegmentation}} &68.3 &61.9 &91.5 &98.2 &81.4 &0.0 &23.3 &65.3 &40.0 &75.5 &87.7 &58.5 &67.8 &65.6 &49.4 \\
{MinkowskiNet\cite{Choy20194DNetworks}} &71.7 &65.4 &91.8 &98.7 &86.2 &0.0 &34.1 &48.9 &62.4 &81.6 &89.8 &47.2 &74.9 &74.4 &58.6 \\
{Stratified Transformer\cite{Lai_2022}}  &78.1 &72.0 &\textbf{96.2} &\textbf{98.7} &85.6 &0.0 &46.1 &60.0 &\textbf{76.8} &\textbf{92.6} &84.5 &77.8 &75.2 &78.1 &\textbf{64.0} \\
{Fast Point Transformer \cite{Park_2022}} &77.9 &70.3 &94.2 &98.0 &86.0 &0.2 &\textbf{53.8} &61.2 &77.3 &81.3 &89.4 &60.1 &72.8 &80.4 &58.9 \\
{CBL} \cite{Tang_2022_CVPR} & 90.6 & 69.4 &93.9&98.4&84.2&0.0&37.0&57.7&71.9&91.7&81.8&77.8&75.6&69.1&62.9 \\\hline
{{PointNet++ }\cite{Qi2017PointNet++:Space}} & 64.1 &55.8 & 88.5 & 94.8 & 74.9 & 0.0 & 25.2& 46.7&39.5&79.8&69.9&63.4&51.4&45.5& 45.8\\ 
\cellcolor[HTML]{E8E8E8}{\textcolor{orange}{+ GeoSpark}}&\cellcolor[HTML]{E8E8E8}69.4 \textcolor{cyan}{(+5.3)} &\cellcolor[HTML]{E8E8E8}  61.1  \textcolor{cyan}{(+5.3)}&\cellcolor[HTML]{E8E8E8} 90.5&\cellcolor[HTML]{E8E8E8}97.8&\cellcolor[HTML]{E8E8E8}80.3&\cellcolor[HTML]{E8E8E8}0.0&\cellcolor[HTML]{E8E8E8} 23.3&\cellcolor[HTML]{E8E8E8}52.7&\cellcolor[HTML]{E8E8E8}59.8&\cellcolor[HTML]{E8E8E8}87.8&\cellcolor[HTML]{E8E8E8}76.6&\cellcolor[HTML]{E8E8E8}69.6&\cellcolor[HTML]{E8E8E8}48.9&\cellcolor[HTML]{E8E8E8}54.0&\cellcolor[HTML]{E8E8E8} 52.5  \\ \hline
{{KPConv}\cite{Thomas2019KPConv:Cloudsb}}  & 70.9 & 65.4 &92.6&97.3&81.4&0.0&16.5&54.5&69.5&90.1&80.2 &74.6& 66.4 &63.7&58.1 \\ 
\cellcolor[HTML]{E8E8E8}\textcolor{orange}{+ GeoSpark} &\cellcolor[HTML]{E8E8E8}72.2 \textcolor{cyan}{(+1.3)} &\cellcolor[HTML]{E8E8E8}65.8 \textcolor{cyan}{(+0.4)}&\cellcolor[HTML]{E8E8E8}  92.4&\cellcolor[HTML]{E8E8E8} 98.4&\cellcolor[HTML]{E8E8E8} 83.3&\cellcolor[HTML]{E8E8E8} 0.0&\cellcolor[HTML]{E8E8E8} 37.7&\cellcolor[HTML]{E8E8E8} 60.2&\cellcolor[HTML]{E8E8E8} 53.2&\cellcolor[HTML]{E8E8E8} 89.3&\cellcolor[HTML]{E8E8E8} 80.6&\cellcolor[HTML]{E8E8E8} 69.2&\cellcolor[HTML]{E8E8E8} 65.5&\cellcolor[HTML]{E8E8E8} 73.0&\cellcolor[HTML]{E8E8E8} 52.0\\  \hline
{{Point Transformer}\cite{Zhao_2021}}  &76.5 &70.4 &94.0 &98.5 &86.3 &0.0 &38.0 &63.4 &74.3 &89.1 &82.4 &74.3 &80.2 &76.0 &59.3 \\ 
\cellcolor[HTML]{E8E8E8}\textcolor{orange}{+ GeoSpark} &\cellcolor[HTML]{E8E8E8}77.7 \textcolor{cyan}{(+1.2)} &\cellcolor[HTML]{E8E8E8} 71.5 \textcolor{cyan}{(+1.1)}  &\cellcolor[HTML]{E8E8E8} 93.7 &\cellcolor[HTML]{E8E8E8} \textbf{98.7}  &\cellcolor[HTML]{E8E8E8} \textbf{86.6} &\cellcolor[HTML]{E8E8E8} 0.2 &\cellcolor[HTML]{E8E8E8} 48.5 &\cellcolor[HTML]{E8E8E8} 60.9 &\cellcolor[HTML]{E8E8E8} 66.4 &\cellcolor[HTML]{E8E8E8} 83.0 &\cellcolor[HTML]{E8E8E8} \textbf{92.7} &\cellcolor[HTML]{E8E8E8} \textbf{82.7} &\cellcolor[HTML]{E8E8E8} 76.2 &\cellcolor[HTML]{E8E8E8} \textbf{81.1} &\cellcolor[HTML]{E8E8E8} 59.3\\
			\bottomrule
	\end{tabular}}
        \vspace{1mm}
	\caption{\textbf{Semantic segmentation results on the S3DIS Area 5.} Consistency improvements are observed after adding GeoSpark to three backbone structures.}
	\label{s3dis_pre_class}
\end{table*}

The network architecture using PointTransformer as the backbone is presented below. For further information on the settings used for other backbones, please refer to the Supplementary Materials. The regularization strength \(\lambda\), was set to 3 for S3DIS and 2 for ScanNetV2 to obtain expressive partitions. Within the global branch, the Superpoints were down-sampled by a ratio of 1/2 after each stage. While the feature dimensions were chosen as [32, 64, 128, 256, 512] for stages 1-5.
Regarding Geometric Downsampling, a size cap of [0.10, 0.20, 0.40, 0.80] m was set for S3DIS, and [0.25, 0.50, 0.75, 1.00] m for ScanNetV2, during stages 1-5.
In the global branch, aggregation was only performed once at each stage. Meanwhile, in the local branch, the depths were set to [1, 2, 2, 6, 2] for stages 1-5, following the approach in \cite{Zhao_2021}. The depth settings were the same for both S3DIS and ScanNetV2 datasets.

\begin{table}[hb]
\small
\centering
\begin{tabular}{l|c|l}
\toprule
\small
\textbf{Method} & Input & \textbf{mIoU} (\%) \\
\hline
PointNet++\cite{Qi2017PointNet++:Space} & point & 53.5 \\
PointCov\cite{Wu2018PointConv:Clouds} & point& 61.0\\
JointPointBased\cite{Hu_2020} & point& 69.2 \\
PointASNL\cite{Yan2020PointASNL:Sampling} & point& 63.5 \\
KPConv\cite{Thomas2019KPConv:Clouds} & point& 69.2 \\
SparseConvNet\cite{Graham_2015} & voxel & 69.3 \\
MinkowskiNet\cite{Choy20194DNetworks} & voxel & 72.2 \\
Fast Point Transformer \cite{Park_2022} & point & 72.1 \\ 
Stratified Transformer\cite{Lai_2022} & point & 74.3 \\  \hline
PointNet++\cite{Qi2017PointNet++:Space} & point & 53.5 \\
\cellcolor[HTML]{E8E8E8}{\textcolor{orange}{+ GeoSpark}} & \cellcolor[HTML]{E8E8E8} point & 63.5 \cellcolor[HTML]{E8E8E8} \textcolor{cyan}{(+10.0)} \\ \hline
KPConv\cite{Thomas2019KPConv:Clouds} &  point & 68.6\\
\cellcolor[HTML]{E8E8E8}{\textcolor{orange}{+ GeoSpark}} & \cellcolor[HTML]{E8E8E8} point & \cellcolor[HTML]{E8E8E8} 71.1 \textcolor{cyan}{(+2.5)} \\ \hline
\textit{Point Transformer}\cite{Zhao_2021} &  point &  70.6 \\
\cellcolor[HTML]{E8E8E8}{\textcolor{orange}{+ GeoSpark}} & \cellcolor[HTML]{E8E8E8} point & \cellcolor[HTML]{E8E8E8} \textbf{74.7} \textcolor{cyan}{(+4.1)} \\ \bottomrule
\end{tabular}
\vspace{1mm}
\caption{\textbf{Semantic segmentation results on ScanNetV2 Validation set.} Adding GeoSpark to the three backbone structures resulted in a noticeable, and the PointTransformer + GeoSpark achieved a higher mIOU than the transformer variation architectures. ranking top on the benchmarks}
\label{scanv2}
\vspace{-3mm}
\end{table}

\paragraph{Implementation details.}
All experiments were conducted using two A100 GPUs. The following presents details of model using Point Transformer as THE backbone. For S3DIS, the model was trained for 30,000 iterations with a batch size of 16. Following previous works \cite{Zhao_2021, Lai_2022}, the input sets were subsampled with a 4mm grid. Area 5 was used for testing while other areas were used for training. The loss weight \(\beta\) was set to 0.1, and the training was carried out using the \textit{AdamW}, optimizer, with a learning rate of 0.004 and a weight decay of 0.02.
As for ScanNetV2, the model was trained for 600 epochs with a batch size of 12. The input sets were subsampled with a 2mm grid, and the \textit{AdamW} optimizer was also used with a learning rate of 0.005 and a weight decay of 0.05. Further implementation details can be found in the Supplementary Materials.

\subsection{Comparisons against State-of-the-arts}

\begin{table*}[t]
    \begin{minipage}{.48\textwidth}
\centering
\begin{tabular}{c|ccc|ccc}\toprule
\textbf{ID} &\textbf{GD} &\textbf{GIA} &\textbf{\(\mathcal{L}_{sup}\)} &{mIoU} &mAcc &OA \\\midrule
I & & & & 72.4 & 81.3 & 90.6\\
II & &\checkmark & & 72.8 & 81.3 & 90.7\\
III & &\checkmark &\checkmark &73.8 & 82.1 & 91.0\\
IV &\checkmark & & & 73.5 & 81.4 & 90.7\\
\cellcolor[HTML]{E8E8E8}VI &\cellcolor[HTML]{E8E8E8}\checkmark &\cellcolor[HTML]{E8E8E8}\checkmark &\cellcolor[HTML]{E8E8E8}\checkmark &\cellcolor[HTML]{E8E8E8}\cellcolor[HTML]{E8E8E8}{74.7} & \cellcolor[HTML]{E8E8E8}{82.7} &\cellcolor[HTML]{E8E8E8}{91.4}\\
\bottomrule
\end{tabular}
\vspace{1mm}
\caption{\textbf{Module ablation study.}. GD: \textit{Geometric Downsampling}. \textbf{GIA}: \textit{Geometry-Informed Aggregation}. \(\mathcal{L}_{sup}\): Superpoint loss. Each module demonstrated its ability to enhance the final results.}
\vspace{2mm}
\label{MODULE}
    \end{minipage}
    \hspace{8mm}
    \begin{minipage}{.45\textwidth}
\begin{tabular}{c|c|ccc}\toprule 
Dataset & SP. \textit{dia} & mIoU & mACC & OA \\\midrule
\multirow{3}{*}{{S3DIS}} & 0.5 & 71.1 & 76.8 & 91.2\\
& \cellcolor[HTML]{E8E8E8} 1 & \cellcolor[HTML]{E8E8E8} {71.5} & \cellcolor[HTML]{E8E8E8}77.3 &\cellcolor[HTML]{E8E8E8} 91.1 \\
&1.5 &70.8 & 76.5 & 90.7 \\ \midrule
\multirow{3}{*}{{ScanNetV2}} &2 & 74.2 & 82.4 & 91.2 \\
& 2.5 & 73.9 & 82.4& 91.2 \\
& \cellcolor[HTML]{E8E8E8}3 & \cellcolor[HTML]{E8E8E8}{74.7} & \cellcolor[HTML]{E8E8E8}{82.7} & \cellcolor[HTML]{E8E8E8} {91.4} \\
\bottomrule
\end{tabular}
\vspace{1mm}
\caption{\textbf{Ablation study of max superpoint diameter setting.} Selecting the appropriate size of superpoint optimize the performance, and the size is tailored to the dataset.}
\vspace{2mm}
\label{partition_cap}
    \end{minipage} \\
    \begin{minipage}{.45\textwidth}
\centering
\begin{tabular}{c|cc|ccc}\toprule
ID &\(k_{local}\) &\(k_{global}\) & mIoU &mAcc &OA \\\midrule
I &8 &4 &72.1 &80.6 &90.6 \\
II &8 &8 &72.5 &81.7 &90.6 \\
III &10 &5 &73.3 &82.0 &90.9 \\
IV &16 &0 &73.5 &82.1 &91.2 \\
V &16 &16 &74.2 &82.4 &91.2 \\
VI &16 &4 &74.4 &82.7 &90.7 \\
\cellcolor[HTML]{E8E8E8}VII &\cellcolor[HTML]{E8E8E8}16 &\cellcolor[HTML]{E8E8E8} 8 &\cellcolor[HTML]{E8E8E8}{74.7} & \cellcolor[HTML]{E8E8E8}{82.7} &\cellcolor[HTML]{E8E8E8} {91.4} \\
\bottomrule
\end{tabular}
\vspace{0.2mm}
\caption{\textbf{Ablation study on local-global balance.} Maintaining a balance between global and local features is crucial for optimizing feature aggregation while minimizing noise introduction.}
\label{balance}
    \end{minipage}
    \hspace{8mm}
    \begin{minipage}{.48\textwidth}
\centering
\begin{tabular}{c|c|ccc}
\toprule
Method &Sample ratio/size &mIoU & mACC & OA \\\midrule
\multirow{2}{*}{FPS} &\(1/4\) &73.8& 82.1& 91.2 \\
&\(1/6\) &72.3 & 81.0 & 90.1 \\ \hline
\multirow{2}{*}{Voxel} &\([~0.10, 0.20, 0.40, 0.80~]\) &74.2& 82.1& 91.2 \\
&\([~0.25, 0.50, 0.75, 1.00~]\) &74.3 & 81.7 & 90.2 \\ \hline
\multirow{3}{*}{GD} &\([~0.06, 0.15, 0.38, 0.90~]\) &74.1& 82.1& 91.2 \\
&\([~0.10, 0.20, 0.40, 0.80~]\) &74.3& 82.1& 91.2 \\
&\cellcolor[HTML]{E8E8E8} \([~0.25, 0.50, 0.75, 1.00~]\) 
&\cellcolor[HTML]{E8E8E8}{74.7} 
&\cellcolor[HTML]{E8E8E8}{82.7} 
&\cellcolor[HTML]{E8E8E8} {91.4}\\
\bottomrule
\end{tabular}
\vspace{0.5mm}
\caption{\textbf{Ablation Study on Downsampling Approach} Geometric Downsampling (GD) is generally superior to voxel sampling and FPS, regardless of the partition size setting. Additionally, fusing-based sampling outperforms FPS.}
\label{ab_samp}
    \end{minipage}
    \vspace{-5mm}
\end{table*}

\paragraph{Quantitative comparisons.} 
We noticed a consistent enhancement in performance across three backbone structures by integrating the GeoSpark plug-in. For the S3DIS dataset, the PointNet backbone exhibited a notable 5.3\% improvement in mIoU matrix, while the KPconv backbone (Rigid Version) showed an increase of 0.4\% in mIoU and 1.3 \% in mACC performance as well. In addition, by adding GeoSpark to the Point Transformer, we achieved an mIoU of 71.5\%, which surpassed the baseline model by 1.1\% and ranked highly on the benchmarks.

\begin{figure*}[t!]
 \centering
 \includegraphics[width=1\textwidth]{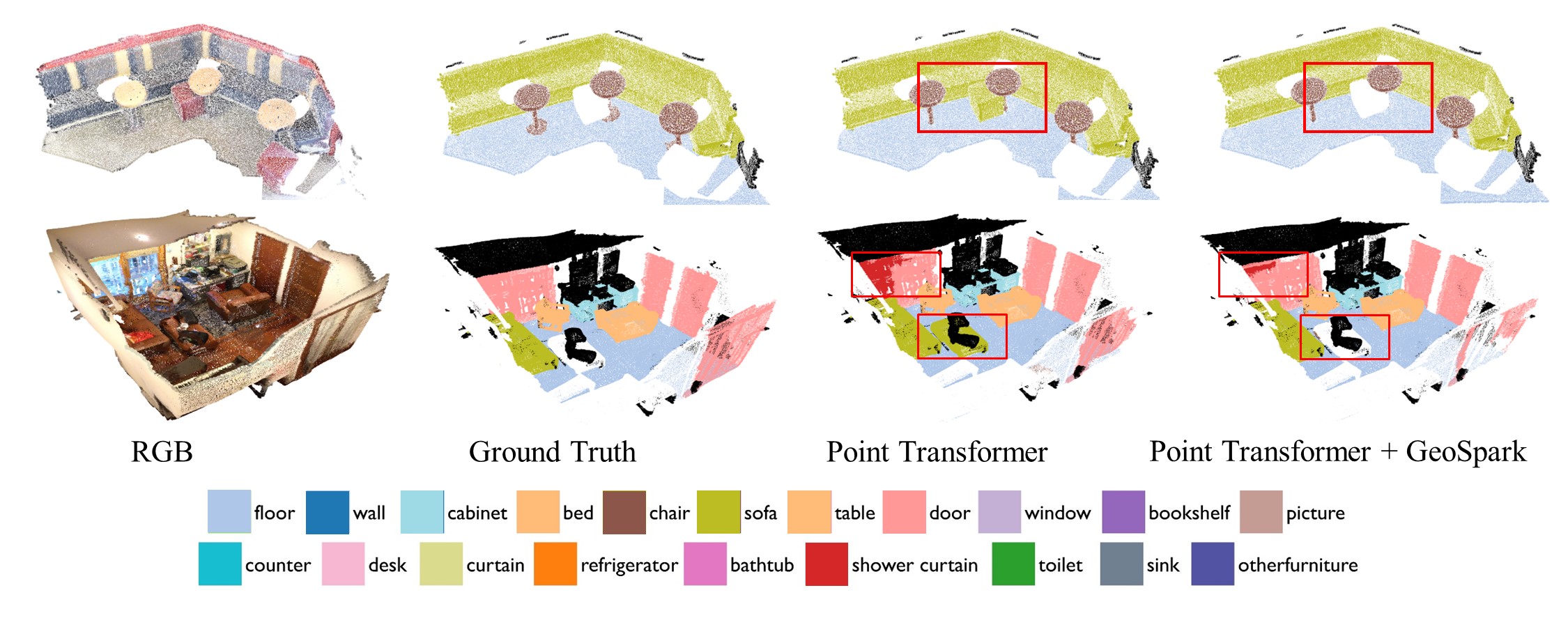}
 
 \caption{\textbf{Visual comparison of the Point Transformer with and without GeoSpark.} GeoSpark captures finer details and produces better predictions for certain objects such as chairs, tables, sofas, and cabinets. The main difference between the prediction results of the Point Transformer and the Point Transformer with GeoSpark was highlighted in a red block}
 \label{fig:visual}
 \centering
 \vspace{0mm}
 \includegraphics[width=1\textwidth]{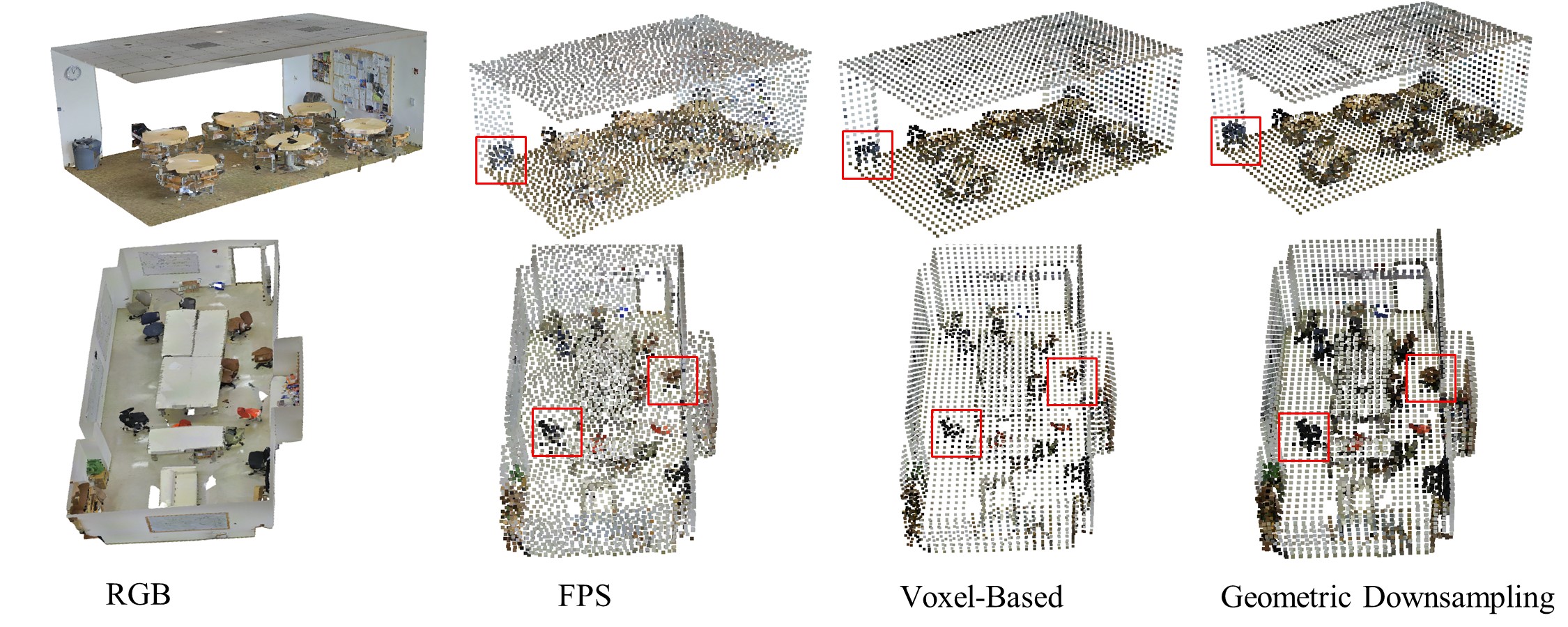}
 \vspace{-6mm}
 \caption{\textbf{Visualization of various sampling methods on the S3DIS dataset}. In comparison to FPS and Voxel-based sampling, Geometric Downsampling preserves key points better, particularly in geometry complex areas. This provides significant benefits, especially for small objects.The red block highlights the preservation of details for small objects.}
 \label{fig:visual_downsamppling}
 \vspace{-5mm}
\end{figure*}

On the ScanNetV2 validation set, integrating GeoSpark to PointNet++ backbone saw a remarkable improvement of 10.0\%, outperforming many other complex designs. This result further demonstrates that modeling long-range features are crucial for accurate large-scale scene understanding. The KPConv backbone demonstrated an increase of 2.5\% in mIoU performance as well. By incorporating Geospark into the Point Transformer, we achieved an mIoU score of 74.7\%, which was a 4.1\% improvement over the baseline model. This approach also surpassed other transformer-based networks, like Fast Point Transformer and Stratified Transformer, indicating the usefulness of geometry guidance in aggregation and sampling modules. 
Our method performed exceptionally well for smaller classes such as chairs, sofas, and boards, as shown in Table \ref{s3dis_pre_class}, highlighting the significance of geometry features in small class segmentation, as shown in Table Further information regarding training time and the number of parameters can be found in the supplementary materials.
\vspace{-3mm}
\paragraph{Visual results}
Figure \ref{fig:visual} presents the visual prediction outcomes of Point Transformer both with and without GeoSpark. Furthermore, Figure \ref{fig:visual_downsamppling} compared our GD approach to other sampling techniques, including FPS and voxel-based sampling.

\subsection{Ablation Study}
A number of controlled experiments were carried out to evaluate the decisions made during the model design process. The ablations were tested on the ScanNetV2 dataset and employed Point Transformer as the backbone. 
\vspace{-4mm}
\paragraph{Module ablation.} For a fair comparison, Point Transformer is retrained with the same experiment setting, including feature dimensions, module depths, and data argumentation. By upgrading the attention module to \textit{GIA}, a substantial improvement is observed. With the help of pseudo-label loss, a 1.4\% gain in mIoU is achieved. This significant improvement confirmed our hypothesis that local aggregation suffers from losing global context and that the \textit{GIA} module is capable of providing meaningful long-range contexts for better aggregation. Changing the sampling strategies also offered a boost in performance, improving the mIoU to 73.5\%. The joining of both modules with the pseudo label loss achieves a final result of 74.7\%. 
\vspace{-3mm}
\paragraph{Local-global balance.}
We then investigate the setting of the number of neighbour points \(k_{local}\), and \(k_{global}\), as shown in Table \ref{balance}. The best results are yielded when the attention module learns from 16 local points and 8 superpoints for feature aggregation. Increasing the number of global points does not provide significant benefits due to the possible introduction of noise. Also, we notice that it is possible to reduce the number of local points, such as (\(k_{local}\) = 10, \(k_{global}\) = 5) set, which achieves similar results as the baseline.
\vspace{-3mm}
\paragraph{Sampling methods.} The "Voxel" approach partitioned input with a voxel grid and uses pooling operations to fuse points. In Table \ref{ab_samp}, we notice that, with the help of geometric contexts, the GD approach indeed significantly improves the results compared to the FPS baseline. Interestingly, we also see an improvement in the voxel-based approaches. A potential reason is that Voxel-based sampling can better preserve geometry features, than that of FPS (as shown in the comparison in Figure \ref{fig:GD}), leading to a boost in performance. With the optimal sampling size setting, GD yields the best results and outperformed its voxel counterparts, demonstrating the benefits of sampling with the geometry-aware approach. 
\vspace{-4mm}

\paragraph{Geometric partition size.} We investigate the different sizes of caps for geometric partition (Table \ref{partition_cap}). For S3DIS, 1.0 m is the optimal size for geometric partitions, while for ScanNetV2, 3.0 m yields the best results, since the data are slightly large. We notice that it is essential to have the partition cap set to be larger than the maximum receptive field at the end of the encoder, to provide extra global contexts while reducing the risk of introducing noise.

\section{Conclusion}
Our work, GeoSpark, aims to utilize explicit geometry guidance to enhance feature learning and downsampling. It leverages easily obtained geometry partition information to guide the feature learning process, resulting in a significant boost to the baseline models. Querying global features from geometric partitions enlarges the receptive fields in a data-dependent way, which helps to capture more expressive features. The geometric partitions also guide the downsampling procedure to drop redundant points while keeping unique points. We hope that this work will draw more attention to integrating geometric partition clues into point cloud understanding, as redundancy is a unique problem of point clouds compared to 2D images.
In the future, we plan to investigate integrating the proposed techniques into other 3D scene understanding tasks, such as object detection and instance segmentation, Additionally, we aim to test the proposed method on outdoor datasets, where long-term feature learning and point redundancy pose significant challenges.
\newpage

\appendix

\section*{Appendix 1: Runtime \& Parameters Analysis}
Table \ref{runtime} presents the training time and a number of parameters on the S3DIS dataset, based on experiments conducted using the Point Transformer backbone, which yielded top-ranking results. The inclusion of a global branch results in an increase in the number of parameters, but it has a manageable effect on training time, as shown in Table \ref{runtime}. This is because the global branch processes points that are typically 3-4 orders of magnitude smaller than the original points (approximately 700-1000 points per S3DIS scan and 900-1200 points on ScanNetV2 scans). The smaller size of global points allows faster processing by the network. Despite the fact that the Geospark+Point Transformer slightly underperforms Stratified Transformer on the S3DIS dataset, we have fewer parameters and shorter training times, demonstrating the design's greater efficiency.

The \textit{geometry partition} module can process 100k points within 2-3 seconds, and the entire S3DIS dataset can be processed in less than an hour on a modern eight-core CPU.

\begin{table}[!htp]\centering
\caption{\textbf{Run time \& No. Parameter Comparison on S3DIS}. Tested on two A100 GPUs with batch size 16. }
\scriptsize
\vspace{2mm}
\begin{tabular}
{lcc}\toprule
\label{runtime}
Methods& Train time(H)$\downarrow$ &No.Para(M) $\downarrow$ \\ \hline
PointTransformer &26 &10.2  \\ 
PointTransformer + GeoSpark &30 \textcolor{purple}{(+15\%)} &15.3 \textcolor{purple}{(+50\%)} \\
Stratified Transformer &100 \textcolor{purple}{(+233\%)} &18.8 \textcolor{purple}{(+84\%)} \\
SparseUNet &42\textcolor{purple}{(+61\%)} &38.0 \textcolor{purple}{(+272\%}\\
\bottomrule
\end{tabular}
\end{table}
\maketitle
\ificcvfinal\thispagestyle{empty}\fi

{\small
\bibliographystyle{ieee_fullname}
\bibliography{egbib}
}

\end{document}